\documentclass{article}

\usepackage{arxiv}

\usepackage[utf8]{inputenc} 
\usepackage[T1]{fontenc}    
\usepackage{lmodern}        
\usepackage{hyperref}       
\usepackage{url}            
\usepackage{booktabs}       
\usepackage{amsfonts}       
\usepackage{nicefrac}       
\usepackage{microtype}      
\usepackage{graphicx}

\title{The Promise and Peril of Artificial Intelligence - ``Violet
Teaming'' Offers a Balanced Path Forward}

\author{
    Alexander J. Titus
    \thanks{Corresponding author -
\href{mailto:publications@theinvivogroup.com}{\nolinkurl{publications@theinvivogroup.com}}}
   \\
    Bioeconomy.XYZ \\
    In Vivo Group \\
  Washington, DC, USA \\
  \texttt{} \\
   \And
    Adam H. Russell
   \\
    Information Sciences Institute \\
    University of Southern California \\
  Los Angeles, CA, USA \\
  \texttt{} \\
  }

\usepackage{longtable,booktabs,array}
\usepackage{calc} 
\usepackage{etoolbox}
\makeatletter
\patchcmd\longtable{\par}{\if@noskipsec\mbox{}\fi\par}{}{}
\makeatother
\IfFileExists{footnotehyper.sty}{\usepackage{footnotehyper}}{\usepackage{footnote}}
\makesavenoteenv{longtable}

\newlength{\cslhangindent}
\newlength{\csllabelwidth}
\newlength{\cslentryspacingunit} 
  {}%
  {\par}
\newenvironment{CSLReferences}[2] 
 {
  \setlength{\parindent}{0pt}
  \ifodd #1
  \let\oldpar\par
  \def\par{\hangindent=\cslhangindent\oldpar}
  \fi
  \setlength{\parskip}{#2\cslentryspacingunit}
 }%
 {}
\usepackage{calc}

\begin{document}
\maketitle

\begin{abstract}
Artificial intelligence (AI) promises immense benefits across sectors,
yet also poses risks from dual-use potentials, biases, and unintended
behaviors. This paper reviews emerging issues with opaque and
uncontrollable AI systems, and proposes an integrative framework called
``violet teaming'' to develop reliable and responsible AI. Violet
teaming combines adversarial vulnerability probing (red teaming) with
solutions for safety and security (blue teaming), while prioritizing
ethics and social benefit. It emerged from AI safety research to manage
risks proactively by design. The paper traces the evolution of red,
blue, and purple teaming toward violet teaming, then discusses applying
violet techniques to address biosecurity risks of AI in biotechnology.
Additional sections review key perspectives across law, ethics,
cybersecurity, macrostrategy, and industry best practices essential for
operationalizing responsible AI through holistic technical and social
considerations. Violet teaming provides both philosophy and method for
steering AI trajectories toward societal good. With conscience and
wisdom, the extraordinary capabilities of AI can enrich humanity. But
without adequate precaution, the risks could prove catastrophic. Violet
teaming aims to empower moral technology for the common welfare.
\end{abstract}


\hypertarget{introduction}{%
\section{Introduction}\label{introduction}}

Artificial intelligence (AI) stands poised to revolutionize every sector
of society, from healthcare (Meenigea and Kolla 2023) to education
(Nguyen et al. 2023), finance (Cao 2022), agriculture (Javaid et al.
2023), transportation (Zheng et al. 2023), communications (Ahammed,
Patgiri, and Nayak 2023), and defense (NSCAI 2021). However, the rapid
pace of advancement in AI over the past decade has concurrently given
rise to valid concerns about dual-use potentials, vulnerabilities,
unintended consequences, and ethical risks that span from financial
fraud to political manipulation, toxic content proliferation, public
safety threats from autonomous systems, and more recently, emerging
dangers like engineering of pathogens (Urbina et al. 2022) or autonomous
weapons enabled by AI (Brundage et al. 2018).

This paper reviews the accelerating landscape of progress in AI
capabilities that underscore the transformative potential of AI across
all facets of public and private life. It surveys risks that have
concurrently emerged from increased reliance on AI systems prone to
unintended behaviors, adversarial exploits, inherent biases, and
opacity. The paper goes on to propose that an integrated framework
called ``violet teaming'' offers a proactive approach to developing AI
that is trustworthy, safe, and socially responsible by design (Aviv
Ovadya 2023) . It traces the conceptual evolution of red, blue, and
purple teaming practices in cybersecurity toward the more recent advent
of violet teaming in AI safety research. To illustrate applied violet
teaming in practice, the paper includes a discussion on methods for
proactively addressing dual-use risks of AI in the high-stakes context
of biotechnology and life sciences research (Alexander J. Titus 2023).

\hypertarget{the-evolution-of-artificial-intelligence-from-theory-to-general-capabilities}{%
\section{The Evolution of Artificial Intelligence: From Theory to
General
Capabilities}\label{the-evolution-of-artificial-intelligence-from-theory-to-general-capabilities}}

The progression of artificial intelligence as a field spans back to
foundational work in the 1950s on mathematical logic, knowledge
representation, search algorithms, theory of computation, and neural
networks. The term ``artificial intelligence'' itself was coined in 1956
at the Dartmouth Conference, which convened pioneering researchers like
John McCarthy, Marvin Minsky, Claude Shannon, and Nathaniel Rochester to
crystallize the new field (McCarthy et al. 2006).~

Influential early systems of this era included the Logic Theorist for
automated theorem proving, the General Problem Solver architecture for
reasoning and search, Dendral for scientific hypothesis generation, and
perceptron networks mimicking neural learning (Newell, Shaw, and Simon
1959; Lindsay et al. 1993; Rosenblatt 1958). However, despite high
hopes, progress stalled after this promising start as the difficulty of
emulating human cognition became apparent. This period from the late
1960s to 1980s became known as the ``AI winter'' as funding dried up.~

But interest was rekindled beginning in the late 1980s and 1990s with
the advent of new statistical and algorithmic approaches like Bayesian
networks, support vector machines, hidden Markov models, and multi-layer
neural network backpropagation. The rise of big data and increased
computing power unlocked new capabilities. Notable milestones in the
modern resurgence include IBM's Deep Blue defeating world chess champion
Garry Kasparov in 1997 using massively parallel search algorithms
(Campbell, Hoane, and Hsu 2002) and the DARPA Grand Challenge spurring
autonomous vehicle development in the 2000s (Buehler, Iagnemma, and
Singh 2009).

But the current era of dramatic breakthroughs emerged in 2012 coinciding
with the revival of deep learning fueled by GPU computing power. Deep
learning refers to neural networks with many layers and hierarchical
feature representation learning capabilities (LeCun, Bengio, and Hinton
2015). Whereas early artificial neural networks contained thousands of
parameters, contemporary state-of-the-art models now utilize hundreds of
billions of parameters (Smith et al. 2022), with the largest exceeding
one trillion parameters (Ren et al. 2023).~

The modern period of AI progress reflects a shift from narrowly focused
applications toward increasingly capable and general systems, especially
in domains like computer vision, natural language processing, robotics,
and gaming. Key inflection points include AlexNet revolutionizing image
recognition with neural networks in 2012, generative adversarial
networks (GANs) for image synthesis in 2014, AlphaGo mastering the game
of Go through reinforcement learning in 2016, and Transformer
architectures like BERT and GPT-3 unlocking order-of-magnitude
performance gains in language tasks starting in 2018 (Krizhevsky,
Sutskever, and Hinton 2012; Goodfellow et al. 2014; Silver et al. 2016;
Vaswani et al. 2017; Brown et al. 2020).

The Transformer enabled attention mechanisms for discerning contextual
relationships in data, replacing recurrence in models like long
short-term memory (LSTMs) models. GPT-3 demonstrated wide linguistic
mastery by pre-training on enormous corpora exceeding one trillion words
(Brown et al., 2020). The size and versatility of models continues to
grow rapidly, with programs like Anthropic's Claude and Google's PaLM
exceeding 500 billion parameters on the path toward artificial general
intelligence (Bubeck et al. 2023). Beyond natural language processing,
areas like computer vision, robotics, and reinforcement learning have
witnessed similar leaps in capability and versatility fueled by scale of
data and models. The pace of advances continues unabated as innovations
build upon each other across all subfields of AI.

\hypertarget{emerging-dual-use-risks-and-vulnerabilities-in-ai-systems}{%
\section{Emerging Dual-Use Risks and Vulnerabilities in AI
Systems}\label{emerging-dual-use-risks-and-vulnerabilities-in-ai-systems}}

The fruits of recent AI advancement are readily visible in
transformative applications spanning autonomous vehicles, personalized
medicine, intelligent infrastructure such as automated management of
data centers, advanced manufacturing, automated cyber defense, and much
more. However, the flip side of increasingly capable AI systems
permeating the real world is that they also expand the potential for
harm via intentional misuse, adversarial exploits, inherent biases, or
unintended behaviors.~

Documented dangers span from financial fraud and social manipulation
enabled by generative AI to cyber attacks on public and private
infrastructure, toxic content proliferation (Pavlopoulos et al. 2020),
embedded biases and discrimination, loss of digital privacy, and
emerging threats associated with autonomous weapons (Klare 2023),
engineered pathogens, or uncontrolled superintelligent systems. Safety
challenges pervade AI subfields including computer vision, natural
language, robotics, and reinforcement learning (Amodei et al. 2016).

Recent examples of damages connected to real-world AI systems include
biased algorithms reinforcing discrimination (Malek 2022) and denying
opportunities (Zeide 2022), generative models spreading misinformation
to influence geopolitics (Ho and Nguyen 2023), ransomware attacks
disrupting critical systems (Aloqaily et al. 2022), unsafe demos of
incomplete capabilities like Meta's Galactica model (Will Douglas Heaven
2022), and fatal accidents involving autonomous vehicles (Koopman and
Fratrik 2019). Unforeseen behaviors arise in part because model
complexity now exceeds human interpretability and controllability.
Opacity exacerbates risks along with accountability gaps. Discriminatory
data baked into training datasets further compounds harm potentials
(Leslie 2019).

While ethics oversight of AI development has expanded, governance
remains fragmented across public and private entities. More
comprehensive solutions are critically needed to promote trustworthy
innovation as rapidly advancing capabilities continue permeating all
facets of life. Without foresight and care, advanced AI could pose
catastrophic risks, underscoring the urgency of multidisciplinary
research toward beneficial AI.

\hypertarget{integrating-red-teaming-blue-teaming-and-ethics-with-violet-teaming}{%
\section{Integrating Red Teaming, Blue Teaming, and Ethics with Violet
Teaming}\label{integrating-red-teaming-blue-teaming-and-ethics-with-violet-teaming}}

Confronting the complex dual-use landscape of AI and managing associated
risks requires reactive and proactive measures. Traditional
cybersecurity paradigms like red teaming and blue teaming provide useful
foundations. Red teaming refers to probing vulnerabilities in a system
as an adversary might to reveal gaps, like penetration testing (Zenko
2015). Blue teaming develops defenses against threats, designing
protections, monitoring, and mitigation (Murdoch and Gse 2014). There is
a growing body of work at large and small companies, at major hacker
conferences such as Black Hat and DEFCON, and across academia to red
team emerging generative AI models (Oremus 2023). While this progress is
welcome by many, there is a need to pair these technological assessments
with an adaptation and design of existing and future models to take into
account sociotechnological ``values'' as well.

Red teaming provides awareness of risks, while blue teaming responds
with solutions. Purple teaming combines both for holistic technological
security assessment (Oakley 2019). However, even these can prove
insufficient as AI systems continuously adapt with retraining and new
data, especially in high-stakes contexts like defense, finance, and
healthcare.

Violet teaming represents an evolution by incorporating consideration of
social benefit directly into design, not just as an add-on. It moves
from reactive to proactive security, building sociotechnical systems
that are robust, safe, and responsible by design (Aviv Ovadya 2023). The
concept emerged in AI safety research grappling with risks of misuse and
unintended behaviors.

\begin{figure}
\hypertarget{violetteaming}{%
\centering
\includegraphics{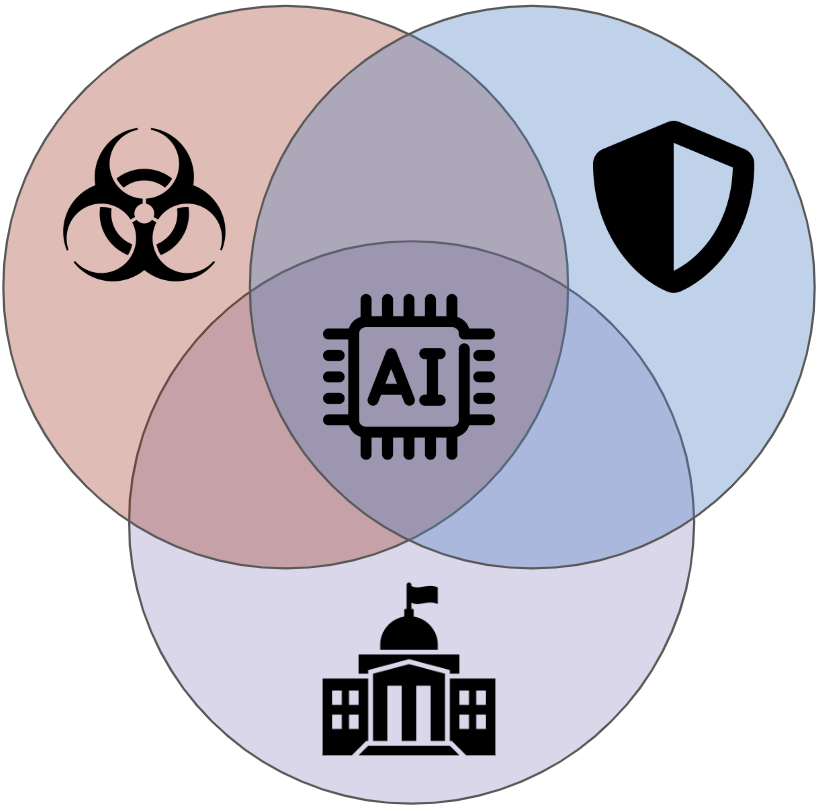}
\caption{Violet teaming is an AI security paradigm that combines the
offensive measures of red teaming and the defensive measures of blue
teaming with a focus on institutional and public benefit
paradigms.}\label{violetteaming}
}
\end{figure}

This new paradigm has been proposed to address emerging biotechnology
risks exacerbated by AI, integrating red team vulnerability assessments
with blue team protections while prioritizing public benefit (Alexander
J. Titus 2023). This proactive approach manages risks by utilizing the
technology itself, not just external oversight. Researchers leverage
techniques like AI to model vulnerabilities and inform technical and
ethical measures inoculating systems against harm. It embeds governance
within the development process rather than as an afterthought.

\hypertarget{bringing-the-social-into-sociotechnical}{%
\subsection{Bringing the Social into
Sociotechnical}\label{bringing-the-social-into-sociotechnical}}

There's rarely much confusion when someone suggests red, blue, or purple
teaming of ``technical systems'' such as cyber networks, large language
models (LLMs), or physical security. While there are obvious and
important nuances between -- as one example - red teaming an LLM to see
if it provides dangerous information vs red teaming the degree to which
an AI system has been trained on biased data, the process of technical
red teaming seems intuitive: test the ``thing'' to see if it does what
it's supposed to do, or not.

Violet teaming recognizes that when it comes to identifying and solving
for AI-induced risks in ways that also advance, rather than hamper,
AI-enabled rewards, then one cannot stay at the purely technical level.~
Instead, violet teaming requires also engaging at the sociotechnical
level, defined as the level where ``technical'' hardware and software
meet, interact with, and reciprocally shape the ``social'' via human
psychology and sociology (Geels 2004; Lee, Dourish, and Mark 2006).~~

The use of the term, sociotechnical, here underscores a true but often
overlooked fact: technologies are the product of, shaped by, embedded
into, and operate via human social systems as much as by acting on the
physical or digital world (Cozzens 1989). If a person creates an AI that
enables themselves to scale up some material or biological product, the
impact is felt at the social level. Once a product is released, an
insight gained, or an outcome achieved, it is in the social world where
people are impacted by those. Thus, the emphasis on incorporating the
sociotechnical as a key feature of violet teaming is intended to
highlight that mitigating AI risks - while enjoying AI's benefits -
requires thinking beyond just the emergent properties of AI's technical
capabilities. It also requires understanding the emergence that occurs
when humans meet AI. There are emergent uses and misuses that new AI
systems create as well as the emergent relationships that AI enables
among machines, humans, and the physical world.

In short, since violet teaming seeks to balance AI's risks and benefits
in ``the real world,'' it seems logical that the real world should not
be ignored or considered irrelevant - but instead should include the
ways people can act with, through, and because of AI's technical
advances. This is critical for helping to design and anticipate AI
systems that afford us the innovations we want and need, while reducing
the chances of outcomes we fear or will regret.

The implications of violet teaming may seem straightforward, but clearly
ask more of us than current red-teaming paradigms of simply
interrogating models to understand what largely purely technical
risk/benefit tradeoffs might be. Depending on the AI design,
capabilities, and outcomes being pursued, violet teaming could mean
incorporating anything from the sociotechnical aspects of bench and wet
lab research organizations to considering population-level behaviors for
things like designing AI to promote participatory democracy.

It is for this reason that the iterative nature of violet teaming is
also emphasized, and that there is no real finish line or point at which
we can simply shrink-wrap AI and forget about it (A. Winfield 2019).
This is what it means for us to be tool-creating apes, capable of
changing ourselves and our systems, where the things we make and use in
today's world invariably lead to a very different world, or future
state.

\hypertarget{research-directions-in-ai-safety-and-violet-teaming}{%
\section{Research Directions in AI Safety and Violet
Teaming}\label{research-directions-in-ai-safety-and-violet-teaming}}

The interdisciplinary field of AI safety focuses on frameworks,
techniques, and guidance for reliable and beneficial systems that avoid
negative consequences (Everitt, Lea, and Hutter 2018). It spans
approaches including robustness, verification, interpretability,
generalization, value alignment, macrostrategy, and policy.

\begin{longtable}[]{@{}
  >{\raggedright\arraybackslash}p{(\columnwidth - 4\tabcolsep) * \real{0.2361}}
  >{\raggedright\arraybackslash}p{(\columnwidth - 4\tabcolsep) * \real{0.5278}}
  >{\raggedright\arraybackslash}p{(\columnwidth - 4\tabcolsep) * \real{0.2361}}@{}}
\caption{Approaches to AI safety research}\tabularnewline
\toprule\noalign{}
\begin{minipage}[b]{\linewidth}\raggedright
Approach
\end{minipage} & \begin{minipage}[b]{\linewidth}\raggedright
Description
\end{minipage} & \begin{minipage}[b]{\linewidth}\raggedright
Reference
\end{minipage} \\
\midrule\noalign{}
\endfirsthead
\toprule\noalign{}
\begin{minipage}[b]{\linewidth}\raggedright
Approach
\end{minipage} & \begin{minipage}[b]{\linewidth}\raggedright
Description
\end{minipage} & \begin{minipage}[b]{\linewidth}\raggedright
Reference
\end{minipage} \\
\midrule\noalign{}
\endhead
\bottomrule\noalign{}
\endlastfoot
Robustness & Guarding against adversarial data, security
vulnerabilities, and spoofing & Goodfellow, Shlens, and Szegedy
(2014) \\
Verification & Formal methods proving correctness of systems and absence
of unintended behaviors & (Katz et al. 2017) \\
Interpretability & Increasing model transparency and explainability for
accountability & (Arrieta et al. 2019) \\
Generalization & Promoting reliability beyond just training data
distributions & (Koh et al. 2020) \\
Value alignment & Developing techniques to align AI goals with human
values and ethics & (Soares and Fallenstein 2015) \\
Macrostrategy & Shaping trajectories of AI and associated technologies
toward beneficial futures & (Mariani 2019) \\
Policy & Developing governance balancing innovation with responsible
oversight & (Jobin, Ienca, and Vayena 2019) \\
\end{longtable}

This research illuminates pathways toward integrative AI systems where
safety is a core feature rather than an afterthought. Violet teaming
aims to unify technical dimensions with ethical and social
considerations under meaningful oversight.

\hypertarget{a-pathway-for-balanced-ai-innovation}{%
\section{A Pathway for Balanced AI
Innovation}\label{a-pathway-for-balanced-ai-innovation}}

External oversight mechanisms like audits, reporting, and review boards
remain indispensable for accountable AI (Raji et al. 2020). But violet
teaming complements these by embedding responsible innovation within the
research and development process itself.~

Violet schemes align red team vulnerability assessments with blue team
solutions to maximize benefits and minimize risks. Initiatives like
DARPA's Guaranteeing AI Robustness Against Deception (GARD) program have
funded pioneering violet methods making models robust by design through
techniques like constrained optimization, value alignment, and recursive
reward modeling (DARPA 2023). Innovative new techniques like
self-destructing models achieve task blocking to frustrate malicious
adaptation of foundation models to harmful uses (Henderson et al. 2023).
Partnerships between industry, government, and civil society can tailor
and scale violet teaming to critical domains like defense, healthcare,
and transportation.

Mainstreaming the violet mindset has potential to steer AI trajectories
toward reliability, security, ethics, and social good by design rather
than as an afterthought. Violet teaming provides both philosophy and
technique for guiding AI toward positive futures.

\hypertarget{violet-teaming-to-address-dual-use-risks-of-ai-in-biotechnology}{%
\section{Violet Teaming to Address Dual-Use Risks of AI in
Biotechnology}\label{violet-teaming-to-address-dual-use-risks-of-ai-in-biotechnology}}

The colossal opportunities of AI must be balanced with risks, as
societal integration accelerates across sectors. The biotechnology
revolution led by CRISPR gene editing has enabled healthcare advances
along with innovations in agriculture, materials, energy, and
environment (Doudna and Charpentier 2014). Democratized bioengineering
also raises concerns of misuse by state adversaries, non-state actors,
or unintended accidents. While regulations aim to prevent misuse,
capabilities are spreading globally (Koblentz and Kiesel 2021).

The advent of AI applied to accelerate biotechnology expands dual-use
risks further. AI is rapidly learning to predict protein folding, design
novel proteins, simulate cellular systems, and synthesize DNA. This
promises immense innovation but could also enable large-scale
engineering of pathogens as bioweapons by more actors (Atlas and Dando
2006).

\hypertarget{approaches-to-start-violet-teaming-ai-in-biotechnology}{%
\subsection{Approaches to start violet teaming AI in
biotechnology}\label{approaches-to-start-violet-teaming-ai-in-biotechnology}}

Violet teaming could be used to constrain generative biotech AI models
by screening hazardous DNA/protein sequences generated during inference
to catch threats before creation (Alexander J. Titus 2023; IARPA 2022).
Rather than just external screening post-design, this embeds internal
checking during generation and utilizes AI capabilities for risk
prevention rather than solely restrictions stifling innovation. The
approach builds collective immunity by inoculating biotechnology with
ethical AI alongside rigorous cybersecurity practices (Alexander J.
Titus, Hamilton, and Holko 2023).

For example, through academic-industry collaborations, one effort could
focus on advancing violet teaming methods for trustworthy AI in
synthetic biology. Open-source software could integrate constrained
optimization and adversarial training to make generative models for
genetic circuit design robust against hazards by screening for risk
factors such as virulence and transmisability during inference. Metrics
could focus on improved reliability on protein engineering tasks while
reducing dual-use potential versus unconstrained models. Extensions
generalizing violet techniques could include probabilistic models and
reinforcement learning. Safety-aware neural architecture search could
identify architectures that balance accuracy and risk, while increasing
accountability through algorithms explaining screening decisions.

\hypertarget{emerging-legislation-focused-on-ai-and-biotechnology}{%
\subsection{Emerging legislation focused on AI and
biotechnology}\label{emerging-legislation-focused-on-ai-and-biotechnology}}

Recently proposed U.S. legislation reflects rising concerns over
dual-use risks of AI intersecting with biotechnology. The Artificial
Intelligence and Biosecurity Risk Assessment Act directs the HHS
Assistant Secretary for Preparedness and Response (ASPR) to assess
whether advancements in AI like open-source models could enable
engineering of dangerous pathogens or bioweapons. It calls for
monitoring global catastrophic biological risks enabled by AI and
incorporating findings into the National Health Security Strategy.
Complementary legislation titled the Strategy for Public Health
Preparedness and Response to Artificial Intelligence Threats Act would
require HHS to develop an integrated preparedness strategy addressing
threats of AI misuse in biotechnology (Edward J. Markey and Ted Budd
2023).

These proposed policies validate concerns about dual-use potentials of
AI and biotech raised by researchers urging governance innovations, like
violet teaming, to mitigate risks while retaining benefits (Alexander J.
Titus 2023). Undue regulation on the use of AI in the life sciences is
likely to have negative economic and national security implications. In
2023, AI-driven drug candidates are entering phase 1 clinical trials and
have demonstrated a significant reduction in time and resources required
to discover drug candidates (Hayden Field 2023). In parallel,
organizations such as the U.S. National Security Commission on
Artificial Intelligence identified biotechnology as a critical domain to
national and economic security (NSCAI 2021).

\hypertarget{violet-teaming-in-support-of-data-driven-policy}{%
\subsection{Violet teaming in support of data-driven
policy}\label{violet-teaming-in-support-of-data-driven-policy}}

Violet teaming's philosophy of pairing vulnerability assessments with
integrated technical and ethical solutions provides a framework for
addressing issues raised in the legislation. For example, HHS could
convene violet teams with AI and biotech expertise to model risks,
stress test systems, and build collective immunity through proactive
measures described in the violet teaming paradigm. Policymakers
recognize the need for applying AI safely in biotechnology, as evidenced
by these proposals. Violet teaming offers principles and methods to
steer innovations toward security and social good that can inform
effective governance.

\hypertarget{macrostrategy-for-responsible-technology-trajectories}{%
\section{Macrostrategy for Responsible Technology
Trajectories}\label{macrostrategy-for-responsible-technology-trajectories}}

Beyond individual applications, the emerging domain of macrostrategy
analyzes how to direct entire technological fields toward beneficial
futures for civilization through prioritized interventions (Dafoe 2018).
With advanced AI, this requires cross-disciplinary insights interfacing
technical factors with political economy, incentives, governance, and
ethics to shape innovation ecosystems holistically. Policy, norms, and
culture that elevate safety, security, and social responsibility as
priorities early can become embedded features enabling positive-sum
outcomes (Mittelstadt 2019). By foregrounding violet teaming goals like
value alignment within research programs, critical infrastructure, and
public discourse, the likelihood of hazards diminishes considerably.
Commitments by technology leaders to uphold ethics help solidify
responsible trajectories and not let undue algorithmic bias harm those
trajectories (O'neil 2017). Avoiding winner-take-all dynamics mitigates
concentration of power over AI that could undermine oversight.
Macrostrategy offers systemic leverage points to tilt uncertain
technosocial systems toward human flourishing rather than dystopia.

\hypertarget{the-path-forward}{%
\section{The Path Forward}\label{the-path-forward}}

This landscape survey across the dimensions of AI safety, ethics,
governance, and macrostrategy aims to synthesize key perspectives and
priorities essential for realizing the promises of AI while navigating
the perils. Operationalizing reliable and responsible AI requires
proactive, holistic integration of technical factors with social
considerations, not just reactive oversight and course correction.
Violet teaming epitomizes this integrative ethos, seeking to steer AI
trajectories toward security without undue bias, accountability, and
service of the common good by design.

The possibilities before us are profound. With conscience and collective
care, the extraordinary capabilities of AI can uplift humanity to new
heights of knowledge, problem-solving, connection, health,
sustainability, creativity, and prosperity for all global citizens. But
without adequate precaution, wisdom, and deliberate efforts to align
design with ethics, the risks could prove catastrophic (Boström 2014).
Our historic opportunities and duties demand the former path. By guiding
AI systems development with moral visions using approaches like violet
teaming, we can aim this most powerful of technologies toward enriching
humanity and our planetary home rather than undermining them. Concerted
action across sectors is needed to mainstream reliability and
responsibility throughout the AI landscape (Jobin, Ienca, and Vayena
2019).

\hypertarget{supplemental-additional-details}{%
\section{Supplemental \& Additional
Details}\label{supplemental-additional-details}}

\hypertarget{broader-initiatives-to-operationalize-responsible-ai}{%
\subsection{Broader Initiatives to Operationalize Responsible
AI}\label{broader-initiatives-to-operationalize-responsible-ai}}

Beyond biotechnology, momentum is building around the world with
initiatives translating responsible AI principles into practice.

\begin{itemize}
\item
  The European Commission proposed regulations introducing mandatory
  risk-based requirements for trustworthy AI design, transparency, and
  governance (Madiega 2021). This elevates violet teaming aims into
  policy.
\item
  Advisory bodies such as the U.S. National Artificial Intelligence
  Advisory Committee (NAIAC) continue advancing best practices across
  the AI life cycle from design to development, testing, and responsible
  deployment. Its May 2023 report highlights the importance of an
  applied governance framework (NAIAC 2023).
\item
  Organizations like the OECD, Stanford's Institute for Human-Centered
  AI, and the Vatican offer guidance on human-centered values critical
  to violet teaming including human dignity, equity, justice,
  sustainability, and common good. Multi-stakeholder collaboration is
  key (Yeung 2020).
\item
  The Alliance for Securing Democracy and similar groups are pioneering
  threat modeling of AI risks across security domains in order to
  strengthen sociotechnical resilience. This exemplifies applied violet
  teaming philosophy (Hagendorff 2020).
\item
  The emerging field of macrostrategy, including scholarship by
  organizations like the Center for Security and Emerging Technology,
  aims to positively shape trajectories of AI and associated
  technologies toward beneficial futures through initiatives at the
  nexus of ethics, governance, and strategic analysis (Schmidt et al.
  2021).
\end{itemize}

This array of efforts underscores growing momentum and appetite for
putting violet ideals into practice across public, private, and civil
society sectors. Our collective future depends on continued progress
toward AI systems that balance advanced capabilities with containment
for the public good.

\hypertarget{human-rights-ethics-and-values-in-ai}{%
\subsection{Human Rights, Ethics, and Values in
AI}\label{human-rights-ethics-and-values-in-ai}}

Promoting human rights, ethics, and justice is central to the violet
teaming vision of responsible AI. Key principles endorsed by
organizations like UNESCO and the Vatican respect for human dignity,
non-discrimination, accessibility and inclusion, privacy, transparency,
accountability, safety and security, environmental well being, and
common good (Jobin, Ienca, and Vayena 2019; UNESCO 2021).

\begin{longtable}[]{@{}
  >{\raggedright\arraybackslash}p{(\columnwidth - 2\tabcolsep) * \real{0.2816}}
  >{\raggedright\arraybackslash}p{(\columnwidth - 2\tabcolsep) * \real{0.7184}}@{}}
\caption{Principles of AI for human rights, ethics, and
values}\tabularnewline
\toprule\noalign{}
\begin{minipage}[b]{\linewidth}\raggedright
Principle
\end{minipage} & \begin{minipage}[b]{\linewidth}\raggedright
Description
\end{minipage} \\
\midrule\noalign{}
\endfirsthead
\toprule\noalign{}
\begin{minipage}[b]{\linewidth}\raggedright
Principle
\end{minipage} & \begin{minipage}[b]{\linewidth}\raggedright
Description
\end{minipage} \\
\midrule\noalign{}
\endhead
\bottomrule\noalign{}
\endlastfoot
Respect for human dignity & Recognizing the irreplaceable value of each
person and not just utility \\
Non-discrimination & Ensuring impartiality free of bias, prejudice or
unfair exclusion \\
Accessibility and inclusion & Enabling equitable participation in the
benefits of AI across all groups \\
Privacy & Safeguarding personal data and individual spheres of
autonomy \\
Transparency & Enabling intelligibility in how AI systems operate to
build trust \\
Accountability & Maintaining clear responsibility and remedy processes
for harms \\
Safety and security & Guaranteeing robustness, reliability and
containment of risks \\
Environmental well being & Honoring human interdependence with the
natural world \\
Common good & Promoting just systems supporting peace, ecology, and
shared prosperity \\
\end{longtable}

Research initiatives seek to develop AI explicitly aligned to such
values in addition to technical objectives (Gabriel 2020). This
underscores the necessity of holistic design encompassing ethics and
human rights alongside utility and performance (Mittelstadt 2019).

\hypertarget{multidisciplinary-perspectives-on-ai-and-society}{%
\subsection{Multidisciplinary Perspectives on AI and
Society}\label{multidisciplinary-perspectives-on-ai-and-society}}

In addition to computer science, many fields offer vital perspectives on
constructing beneficial versus detrimental futures with AI:

\begin{itemize}
\item
  Philosophy investigates ethics of emerging technologies through lenses
  like utilitarianism, deontology, virtue ethics, and justice (A. F.
  Winfield et al. 2019)
\item
  Psychology examines cognition, biases, decision-making, and human
  needs essential for value alignment and human compatibility
\item
  Organization science analyzes institutional contexts enabling
  responsible innovation or vulnerability based on dynamics like
  incentives, culture, and leadership
\item
  Anthropology provides cultural lenses to assess AI impacts on social
  groups and meanings vital to human thriving
\item
  Political science weighs governance regimes and policies shaping AI
  for the public interest versus excess consolidation of power and
  control (Peters 2022)
\item
  Economics furnishes models of incentive structures, market dynamics,
  and valuation assumptions guiding AI developments with distributional
  consequences~
\item
  Sociology investigates collective social phenomena and change
  associated with AI through historical contexts
\item
  Criminology applies risk and prevention frameworks to malicious uses
  of AI
\item
  Communications studies disinformation ecosystems propagated through AI
  (Broniatowski et al. 2018)
\item
  Design disciplines offer human-centered methods balancing values
  amidst complexity and constraints
\item
  Biological perspectives consider AI vis-a-vis human cognition,
  evolution, and neuroscience
\end{itemize}

Synthesizing insights across these diverse fields alongside computing is
crucial for holistic violet teaming and wise co-evolution of humanity
with technology.

\hypertarget{law-policy-and-responsible-ai-governance}{%
\subsection{Law, Policy, and Responsible AI
Governance}\label{law-policy-and-responsible-ai-governance}}

Alongside research, the policy domain is vital for institutionalizing
responsible practices. Organizations like the OECD, European Commission,
and US government have put forward AI governance frameworks centered on
ethical purpose, transparency, accountability, robustness, and oversight
(Whittaker et al. 2018).

Key policy directions include (Fjeld et al. 2020):

\begin{itemize}
\item
  Mandating algorithmic impact assessments and risk mitigation processes
  calibrated to application risks
\item
  Promoting public oversight through mechanisms like algorithmic
  auditing to assess fairness, accuracy, and security (Raji and
  Buolamwini 2019)
\item
  Incentivizing safety engineering and enabling third-party validation
  to reduce vulnerabilities
\item
  Institutionalizing whistleblowing and consumer protection channels to
  identify and remedy harms (Goodman and Trehu 2023)
\item
  Requiring transparency for certain public sector uses and
  business-to-consumer services to increase intelligibility
\item
  Building capacity and public literacy to participate meaningfully in
  AI discourse and systems shaping society (Floridi et al. 2020)
\item
  Supporting interdisciplinary research on trustworthy AI spanning
  technical and social dimensions
\item
  Cultivating organizational cultures valuing ethics, diversity, and
  human centeredness~
\item
  Investing in digital infrastructure and platforms designed for
  collective well being from the start
\end{itemize}

Multifaceted policy, legal, and regulatory mixes tailored to context are
needed rather than single silver bullets. But the key is evolving
governance to guide AI in line with democratic values.

\hypertarget{industry-practice-and-applications-of-trustworthy-ai}{%
\subsection{Industry Practice and Applications of Trustworthy
AI}\label{industry-practice-and-applications-of-trustworthy-ai}}

Technology firms and industry research consortia are also advancing
practices for reliable and responsible AI:

\begin{itemize}
\item
  Rigorous testing protocols assess models across metrics of safety,
  security, fairness, and accountability before real-world deployment.
  Adversarial testing probes model robustness (Ali et al. 2023).
\item
  Techniques like dataset tagging, noise injection, and constraints
  prevent embedding and propagating biases that could compound
  discrimination (Mehrabi et al. 2022).~~
\item
  Granular documentation details data provenance, assumptions,
  architecture, and performance to enable auditing. Version histories
  support reproducibility (Mitchell et al. 2019).
\item
  Quantifying uncertainties provides calibrated confidence to guide
  human judgment in model integration.
\item
  Monitoring systems coupled with human oversight mechanisms assess
  models post-deployment to detect harms or deviations. Feedback informs
  updates (Whittlestone et al. 2019).
\item
  Design thinking synthesizes technical capabilities with holistic needs
  and values of communities affected (Dignum 2017).
\item
  Stakeholder participation mechanisms foster engagement between
  developers, users, and impacted groups.
\item
  Bug bounties and red team exercises incentivize external researchers
  to find flaws, enabling correction before exploitation (Brundage et
  al. 2020).
\end{itemize}

Partnerships across industry, academia, and civil society combine
strengths in building wise governance.

\hypertarget{cybersecurity-and-adversarial-robustness}{%
\subsection{Cybersecurity and Adversarial
Robustness}\label{cybersecurity-and-adversarial-robustness}}

As AI permeates infrastructure and services, cybersecurity is crucial to
ensure resilience against bad actors seeking to manipulate systems for
harm. Core approaches include (Biggio and Roli 2018):

\begin{itemize}
\item
  Adversarial machine learning hardens models against malicious inputs
  designed to cause misclassification, misdirection, and system
  compromise (Goodfellow, Shlens, and Szegedy 2014; Szegedy et al.
  2013).~~
\item
  Differential privacy, homomorphic encryption, secure multi-party
  computation and cryptographic methods safeguard sensitive user data
  (Dwork and Roth 2013; Alexander J. Titus et al. 2018).
\item
  Formal verification mathematically proves system behaviors align to
  specifications under conditions (Katz et al. 2017).
\item
  Software engineering practices like code reviews, penetration testing,
  and building security into the development life cycle.~~
\item
  Monitoring, logging, and anomaly detection surface attacks along with
  system risks and failures to inform mitigation (Chandola, Banerjee,
  and Kumar 2009).
\item
  Cyber deception, involving setting traps to detect, deflect, and
  counter exploits through techniques like honeypots mimicking systems
  that lure attackers. Robust cybersecurity protections integrated with
  violet teaming principles and oversight are imperative as AI-enabled
  technologies are entrusted with sensitive roles (Wang and Lu 2018).
\end{itemize}

\hypertarget{references}{%
\section*{References}\label{references}}
\addcontentsline{toc}{section}{References}

\hypertarget{refs}{}
\begin{CSLReferences}{1}{0}
\leavevmode\vadjust pre{\hypertarget{ref-ahammed2023}{}}%
Ahammed, Tareq B., Ripon Patgiri, and Sabuzima Nayak. 2023. {``A Vision
on the Artificial Intelligence for 6G Communication.''} \emph{ICT
Express} 9 (2): 197--210.
https://doi.org/\url{https://doi.org/10.1016/j.icte.2022.05.005}.

\leavevmode\vadjust pre{\hypertarget{ref-ali2023}{}}%
Ali, Sajid, Tamer Abuhmed, Shaker El-Sappagh, Khan Muhammad, Jose M.
Alonso-Moral, Roberto Confalonieri, Riccardo Guidotti, Javier Del Ser,
Natalia Díaz-Rodríguez, and Francisco Herrera. 2023. {``Explainable
Artificial Intelligence (XAI): What We Know and What Is Left to Attain
Trustworthy Artificial Intelligence.''} \emph{Information Fusion} 99
(November): 101805. \url{https://doi.org/10.1016/j.inffus.2023.101805}.

\leavevmode\vadjust pre{\hypertarget{ref-aloqaily2022}{}}%
Aloqaily, Moayad, Salil Kanhere, Paolo Bellavista, and Michele Nogueira.
2022. {``Special Issue on Cybersecurity Management in the Era of AI.''}
\emph{Journal of Network and Systems Management} 30 (3).
\url{https://doi.org/10.1007/s10922-022-09659-3}.

\leavevmode\vadjust pre{\hypertarget{ref-amodei2016}{}}%
Amodei, Dario, Chris Olah, Jacob Steinhardt, Paul Christiano, John
Schulman, and Dan Mané. 2016. {``Concrete Problems in AI Safety.''}
\url{https://doi.org/10.48550/ARXIV.1606.06565}.

\leavevmode\vadjust pre{\hypertarget{ref-arrieta2019}{}}%
Arrieta, Alejandro Barredo, Natalia Díaz-Rodríguez, Javier Del Ser,
Adrien Bennetot, Siham Tabik, Alberto Barbado, Salvador García, et al.
2019. {``Explainable Artificial Intelligence (XAI): Concepts,
Taxonomies, Opportunities and Challenges Toward Responsible AI.''}
\url{https://doi.org/10.48550/ARXIV.1910.10045}.

\leavevmode\vadjust pre{\hypertarget{ref-atlas2006}{}}%
Atlas, Ronald M., and Malcolm Dando. 2006. {``The Dual-Use Dilemma for
the Life Sciences: Perspectives, Conundrums, and Global Solutions.''}
\emph{Biosecurity and Bioterrorism: Biodefense Strategy, Practice, and
Science} 4 (3): 276--86. \url{https://doi.org/10.1089/bsp.2006.4.276}.

\leavevmode\vadjust pre{\hypertarget{ref-avivovadya2023}{}}%
Aviv Ovadya. 2023. {``Red Teaming Improved GPT-4. Violet Teaming Goes
Even Further.''}
\url{https://www.wired.com/story/red-teaming-gpt-4-was-valuable-violet-teaming-will-make-it-better/}.

\leavevmode\vadjust pre{\hypertarget{ref-biggio2018}{}}%
Biggio, Battista, and Fabio Roli. 2018. {``Wild Patterns: Ten Years
After the Rise of Adversarial Machine Learning.''} \emph{Pattern
Recognition} 84 (December): 317--31.
\url{https://doi.org/10.1016/j.patcog.2018.07.023}.

\leavevmode\vadjust pre{\hypertarget{ref-bostruxf6m2014}{}}%
Boström, Nick. 2014. {``Superintelligence: Paths, Dangers,
Strategies.''} \emph{Superintelligence: Paths, Dangers, Strategies}.

\leavevmode\vadjust pre{\hypertarget{ref-broniatowski2018}{}}%
Broniatowski, David A., Amelia M. Jamison, SiHua Qi, Lulwah AlKulaib,
Tao Chen, Adrian Benton, Sandra C. Quinn, and Mark Dredze. 2018.
{``Weaponized Health Communication: Twitter Bots and Russian Trolls
Amplify the Vaccine Debate.''} \emph{American Journal of Public Health}
108 (10): 1378--84. \url{https://doi.org/10.2105/AJPH.2018.304567}.

\leavevmode\vadjust pre{\hypertarget{ref-brown2020}{}}%
Brown, Tom B., Benjamin Mann, Nick Ryder, Melanie Subbiah, Jared Kaplan,
Prafulla Dhariwal, Arvind Neelakantan, et al. 2020. {``Language Models
Are Few-Shot Learners.''}
\url{https://doi.org/10.48550/ARXIV.2005.14165}.

\leavevmode\vadjust pre{\hypertarget{ref-brundage2018}{}}%
Brundage, Miles, Shahar Avin, Jack Clark, Helen Toner, Peter Eckersley,
Ben Garfinkel, Allan Dafoe, et al. 2018. {``The Malicious Use of
Artificial Intelligence: Forecasting, Prevention, and Mitigation.''}
\url{https://doi.org/10.48550/ARXIV.1802.07228}.

\leavevmode\vadjust pre{\hypertarget{ref-brundage2020}{}}%
Brundage, Miles, Shahar Avin, Jasmine Wang, Haydn Belfield, Gretchen
Krueger, Gillian Hadfield, Heidy Khlaaf, et al. 2020. {``Toward
Trustworthy AI Development: Mechanisms for Supporting Verifiable
Claims.''} \url{https://doi.org/10.48550/ARXIV.2004.07213}.

\leavevmode\vadjust pre{\hypertarget{ref-bubeck2023}{}}%
Bubeck, Sébastien, Varun Chandrasekaran, Ronen Eldan, Johannes Gehrke,
Eric Horvitz, Ece Kamar, Peter Lee, et al. 2023. {``Sparks of Artificial
General Intelligence: Early Experiments with GPT-4.''}

\leavevmode\vadjust pre{\hypertarget{ref-thedarp2009}{}}%
Buehler, Martin, Karl Iagnemma, and Sanjiv Singh, eds. 2009. \emph{The
DARPA Urban Challenge}. Springer Berlin Heidelberg.
\url{https://doi.org/10.1007/978-3-642-03991-1}.

\leavevmode\vadjust pre{\hypertarget{ref-campbell2002}{}}%
Campbell, Murray, A.Joseph Hoane, and Feng-hsiung Hsu. 2002. {``Deep
Blue.''} \emph{Artificial Intelligence} 134 (1-2): 57--83.
\url{https://doi.org/10.1016/S0004-3702(01)00129-1}.

\leavevmode\vadjust pre{\hypertarget{ref-cao2022}{}}%
Cao, Longbing. 2022. {``AI in Finance: Challenges, Techniques, and
Opportunities.''} \emph{ACM Comput. Surv.} 55 (3).
\url{https://doi.org/10.1145/3502289}.

\leavevmode\vadjust pre{\hypertarget{ref-chandola2009}{}}%
Chandola, Varun, Arindam Banerjee, and Vipin Kumar. 2009. {``Anomaly
Detection: A Survey.''} \emph{ACM Computing Surveys} 41 (3): 1--58.
\url{https://doi.org/10.1145/1541880.1541882}.

\leavevmode\vadjust pre{\hypertarget{ref-cozzens1989}{}}%
Cozzens, Susan E. 1989. {``The Social Construction of Technological
Systems: New Directions in the Sociology and History of Technology.''}

\leavevmode\vadjust pre{\hypertarget{ref-dafoe2018}{}}%
Dafoe, Allan. 2018. {``AI Governance: A Research Agenda.''}
\emph{Governance of AI Program, Future of Humanity Institute, University
of Oxford: Oxford, UK} 1442: 1443.

\leavevmode\vadjust pre{\hypertarget{ref-darpa2023}{}}%
DARPA. 2023. {``Guaranteeing AI Robustness Against Deception (GARD).''}

\leavevmode\vadjust pre{\hypertarget{ref-dignum2017}{}}%
Dignum, Virginia. 2017. {``Responsible Artificial Intelligence:
Designing AI for Human Values.''}

\leavevmode\vadjust pre{\hypertarget{ref-doudna2014}{}}%
Doudna, Jennifer A., and Emmanuelle Charpentier. 2014. {``The New
Frontier of Genome Engineering with CRISPR-Cas9.''} \emph{Science} 346
(6213): 1258096. \url{https://doi.org/10.1126/science.1258096}.

\leavevmode\vadjust pre{\hypertarget{ref-dwork2013}{}}%
Dwork, Cynthia, and Aaron Roth. 2013. {``The Algorithmic Foundations of
Differential Privacy.''} \emph{Foundations and Trends® in Theoretical
Computer Science} 9 (3-4): 211--407.
\url{https://doi.org/10.1561/0400000042}.

\leavevmode\vadjust pre{\hypertarget{ref-edwardj.markey2023}{}}%
Edward J. Markey, and Ted Budd. 2023. {``SENS. MARKEY, BUDD ANNOUNCE
LEGISLATION TO ASSESS HEALTH SECURITY RISKS OF AI.''}
\url{https://www.markey.senate.gov/news/press-releases/sens-markey-budd-announce-legislation-to-assess-health-security-risks-of-ai}.

\leavevmode\vadjust pre{\hypertarget{ref-everitt2018}{}}%
Everitt, Tom, Gary Lea, and Marcus Hutter. 2018. {``AGI Safety
Literature Review.''} \url{https://doi.org/10.48550/ARXIV.1805.01109}.

\leavevmode\vadjust pre{\hypertarget{ref-fjeld2020}{}}%
Fjeld, Jessica, Nele Achten, Hannah Hilligoss, Adam Nagy, and Madhulika
Srikumar. 2020. {``Principled Artificial Intelligence: Mapping Consensus
in Ethical and Rights-Based Approaches to Principles for AI.''}
\emph{SSRN Electronic Journal}.
\url{https://doi.org/10.2139/ssrn.3518482}.

\leavevmode\vadjust pre{\hypertarget{ref-floridi2020}{}}%
Floridi, Luciano, Josh Cowls, Thomas C. King, and Mariarosaria Taddeo.
2020. {``How to Design AI for Social Good: Seven Essential Factors.''}
\emph{Science and Engineering Ethics} 26 (3): 1771--96.
\url{https://doi.org/10.1007/s11948-020-00213-5}.

\leavevmode\vadjust pre{\hypertarget{ref-gabriel2020}{}}%
Gabriel, Iason. 2020. {``Artificial Intelligence, Values, and
Alignment.''} \emph{Minds and Machines} 30 (3): 411--37.
\url{https://doi.org/10.1007/s11023-020-09539-2}.

\leavevmode\vadjust pre{\hypertarget{ref-geels2004}{}}%
Geels, Frank W. 2004. {``From Sectoral Systems of Innovation to
Socio-Technical Systems: Insights about Dynamics and Change from
Sociology and Institutional Theory.''} \emph{Research Policy} 33 (6-7):
897920.

\leavevmode\vadjust pre{\hypertarget{ref-goodfellow2014}{}}%
Goodfellow, Ian J., Jean Pouget-Abadie, Mehdi Mirza, Bing Xu, David
Warde-Farley, Sherjil Ozair, Aaron Courville, and Yoshua Bengio. 2014.
{``Generative Adversarial Networks.''}
\url{https://doi.org/10.48550/ARXIV.1406.2661}.

\leavevmode\vadjust pre{\hypertarget{ref-goodfellow2014a}{}}%
Goodfellow, Ian J., Jonathon Shlens, and Christian Szegedy. 2014.
{``Explaining and Harnessing Adversarial Examples.''}
\url{https://doi.org/10.48550/ARXIV.1412.6572}.

\leavevmode\vadjust pre{\hypertarget{ref-goodman2023}{}}%
Goodman, Ellen P, and Julia Trehu. 2023. {``ALGORITHMIC AUDITING:
CHASING AI ACCOUNTABILITY.''} \emph{Santa Clara High Technology Law
Journal} 39 (3): 289.

\leavevmode\vadjust pre{\hypertarget{ref-hagendorff2020}{}}%
Hagendorff, Thilo. 2020. {``The Ethics of AI Ethics: An Evaluation of
Guidelines.''} \emph{Minds and Machines} 30 (1): 99--120.
\url{https://doi.org/10.1007/s11023-020-09517-8}.

\leavevmode\vadjust pre{\hypertarget{ref-haydenfield2023}{}}%
Hayden Field. 2023. {``The First Fully a.i.-Generated Drug Enters
Clinical Trials in Human Patients.''}
\url{https://www.cnbc.com/2023/06/29/ai-generated-drug-begins-clinical-trials-in-human-patients.html}.

\leavevmode\vadjust pre{\hypertarget{ref-henderson2023}{}}%
Henderson, Peter, Eric Mitchell, Christopher D. Manning, Dan Jurafsky,
and Chelsea Finn. 2023. {``Self-Destructing Models: Increasing the Costs
of Harmful Dual Uses of Foundation Models.''}
\url{https://doi.org/10.48550/ARXIV.2211.14946}.

\leavevmode\vadjust pre{\hypertarget{ref-ho2023}{}}%
Ho, Manh-Tung, and Hong-Kong T. Nguyen. 2023. {``Artificial Intelligence
as the New Fire and Its Geopolitics.''} \emph{AI \& SOCIETY}, May.
\url{https://doi.org/10.1007/s00146-023-01678-1}.

\leavevmode\vadjust pre{\hypertarget{ref-iarpa2022}{}}%
IARPA. 2022. {``FUNCTIONAL GENOMIC AND COMPUTATIONAL ASSESSMENT OF
THREATS (FUN GCAT).''}
\url{https://www.iarpa.gov/research-programs/fun-gcat}.

\leavevmode\vadjust pre{\hypertarget{ref-javaid2023}{}}%
Javaid, Mohd, Abid Haleem, Ibrahim Haleem Khan, and Rajiv Suman. 2023.
{``Understanding the Potential Applications of Artificial Intelligence
in Agriculture Sector.''} \emph{Advanced Agrochem} 2 (1): 15--30.
https://doi.org/\url{https://doi.org/10.1016/j.aac.2022.10.001}.

\leavevmode\vadjust pre{\hypertarget{ref-jobin2019}{}}%
Jobin, Anna, Marcello Ienca, and Effy Vayena. 2019. {``The Global
Landscape of AI Ethics Guidelines.''} \emph{Nature Machine Intelligence}
1 (9): 389--99. \url{https://doi.org/10.1038/s42256-019-0088-2}.

\leavevmode\vadjust pre{\hypertarget{ref-katz2017}{}}%
Katz, Guy, Clark Barrett, David L. Dill, Kyle Julian, and Mykel J.
Kochenderfer. 2017. {``Reluplex: An Efficient SMT Solver for Verifying
Deep Neural Networks.''} In, edited by Rupak Majumdar and Viktor Kunčak,
10426:97--117. Cham: Springer International Publishing.
\url{https://link.springer.com/10.1007/978-3-319-63387-9_5}.

\leavevmode\vadjust pre{\hypertarget{ref-klare2023}{}}%
Klare, Michael. 2023. {``Dueling Views on AI, Autonomous Weapons.''}
\emph{Arms Control Today} 53 (3): 3334.

\leavevmode\vadjust pre{\hypertarget{ref-koblentz2021}{}}%
Koblentz, Gregory D., and Stevie Kiesel. 2021. {``The COVID-19 Pandemic:
Catalyst or Complication for Bioterrorism?''} \emph{Studies in Conflict
\& Terrorism}, July, 1--27.
\url{https://doi.org/10.1080/1057610X.2021.1944023}.

\leavevmode\vadjust pre{\hypertarget{ref-koh2020}{}}%
Koh, Pang Wei, Thao Nguyen, Yew Siang Tang, Stephen Mussmann, Emma
Pierson, Been Kim, and Percy Liang. 2020. {``Concept Bottleneck
Models.''} \url{https://doi.org/10.48550/ARXIV.2007.04612}.

\leavevmode\vadjust pre{\hypertarget{ref-koopman2019}{}}%
Koopman, Philip, and Frank Fratrik. 2019. {``How Many Operational Design
Domains, Objects, and Events?''} \emph{Safeai@ Aaai} 4.

\leavevmode\vadjust pre{\hypertarget{ref-krizhevsky2012}{}}%
Krizhevsky, Alex, Ilya Sutskever, and Geoffrey E Hinton. 2012.
{``ImageNet Classification with Deep Convolutional Neural Networks.''}
In, edited by F. Pereira, C. J. Burges, L. Bottou, and K. Q. Weinberger.
Vol. 25. Curran Associates, Inc.
\url{https://proceedings.neurips.cc/paper_files/paper/2012/file/c399862d3b9d6b76c8436e924a68c45b-Paper.pdf}.

\leavevmode\vadjust pre{\hypertarget{ref-lecun2015}{}}%
LeCun, Yann, Yoshua Bengio, and Geoffrey Hinton. 2015. {``Deep
Learning.''} \emph{Nature} 521 (7553): 436--44.
\url{https://doi.org/10.1038/nature14539}.

\leavevmode\vadjust pre{\hypertarget{ref-lee2006}{}}%
Lee, Charlotte P., Paul Dourish, and Gloria Mark. 2006. {``CSCW06:
Computer Supported Cooperative Work.''} In, 483--92. Banff Alberta
Canada: ACM. \url{https://doi.org/10.1145/1180875.1180950}.

\leavevmode\vadjust pre{\hypertarget{ref-leslie2019}{}}%
Leslie, David. 2019. {``Understanding Artificial Intelligence Ethics and
Safety: A Guide for the Responsible Design and Implementation of AI
Systems in the Public Sector.''}
\url{https://doi.org/10.5281/ZENODO.3240529}.

\leavevmode\vadjust pre{\hypertarget{ref-lindsay1993}{}}%
Lindsay, Robert K., Bruce G. Buchanan, Edward A. Feigenbaum, and Joshua
Lederberg. 1993. {``DENDRAL: A Case Study of the First Expert System for
Scientific Hypothesis Formation.''} \emph{Artificial Intelligence} 61
(2): 209--61. \url{https://doi.org/10.1016/0004-3702(93)90068-M}.

\leavevmode\vadjust pre{\hypertarget{ref-madiega2021}{}}%
Madiega, Tambiama André. 2021. {``Artificial Intelligence Act.''}
\emph{European Parliament: European Parliamentary Research Service}.

\leavevmode\vadjust pre{\hypertarget{ref-malek2022}{}}%
Malek, Md. Abdul. 2022. {``Criminal Courts{'} Artificial Intelligence:
The Way It Reinforces Bias and Discrimination.''} \emph{AI and Ethics} 2
(1): 233--45. \url{https://doi.org/10.1007/s43681-022-00137-9}.

\leavevmode\vadjust pre{\hypertarget{ref-mariani2019}{}}%
Mariani, Stefano. 2019. {``Coordination in Socio-Technical Systems:
Where Are We Now? Where Do We Go Next?''} \emph{Science of Computer
Programming} 184 (October): 102317.
\url{https://doi.org/10.1016/j.scico.2019.102317}.

\leavevmode\vadjust pre{\hypertarget{ref-mccarthy2006}{}}%
McCarthy, John, Marvin L. Minsky, Nathaniel Rochester, and Claude E.
Shannon. 2006. {``A Proposal for the Dartmouth Summer Research Project
on Artificial Intelligence, August 31, 1955.''} \emph{AI Magazine} 27
(4): 12. \url{https://doi.org/10.1609/aimag.v27i4.1904}.

\leavevmode\vadjust pre{\hypertarget{ref-meenigea2023}{}}%
Meenigea, Niharikareddy, and Venkata Ravi Kiran Kolla. 2023.
{``Exploring the Current Landscape of Artificial Intelligence in
Healthcare.''} \emph{International Journal of Sustainable Development in
Computing Science} 5 (1).
\url{https://www.ijsdcs.com/index.php/ijsdcs/article/view/285}.

\leavevmode\vadjust pre{\hypertarget{ref-mehrabi2022}{}}%
Mehrabi, Ninareh, Fred Morstatter, Nripsuta Saxena, Kristina Lerman, and
Aram Galstyan. 2022. {``A Survey on Bias and Fairness in Machine
Learning.''} \emph{ACM Computing Surveys} 54 (6): 1--35.
\url{https://doi.org/10.1145/3457607}.

\leavevmode\vadjust pre{\hypertarget{ref-mitchell2019}{}}%
Mitchell, Margaret, Simone Wu, Andrew Zaldivar, Parker Barnes, Lucy
Vasserman, Ben Hutchinson, Elena Spitzer, Inioluwa Deborah Raji, and
Timnit Gebru. 2019. {``FAT* '19: Conference on Fairness, Accountability,
and Transparency.''} In, 220--29. Atlanta GA USA: ACM.
\url{https://doi.org/10.1145/3287560.3287596}.

\leavevmode\vadjust pre{\hypertarget{ref-mittelstadt2019}{}}%
Mittelstadt, Brent. 2019. {``Principles Alone Cannot Guarantee Ethical
AI.''} \emph{Nature Machine Intelligence} 1 (11): 501--7.
\url{https://doi.org/10.1038/s42256-019-0114-4}.

\leavevmode\vadjust pre{\hypertarget{ref-murdoch2014}{}}%
Murdoch, D. W., and D. M. Gse. 2014. \emph{Blue Team Handbook: Incident
Response Edition: A Condensed Field Guide for the Cyber Security
Incident Responder}. Createspace Independent Publishing Platform.
\url{https://books.google.com/books?id=1f7doQEACAAJ}.

\leavevmode\vadjust pre{\hypertarget{ref-naiac2023}{}}%
NAIAC. 2023. {``National Artificial Intelligence Advisory Committee
(NAIAC) Year 1 Report.''}
\url{https://www.ai.gov/wp-content/uploads/2023/05/NAIAC-Report-Year1.pdf}.

\leavevmode\vadjust pre{\hypertarget{ref-newell1959}{}}%
Newell, Allen, John C Shaw, and Herbert A Simon. 1959. {``Report on a
General Problem Solving Program.''} In, 256:64. Pittsburgh, PA.

\leavevmode\vadjust pre{\hypertarget{ref-nguyen2023}{}}%
Nguyen, Andy, Ha Ngan Ngo, Yvonne Hong, Belle Dang, and Bich-Phuong Thi
Nguyen. 2023. {``Ethical Principles for Artificial Intelligence in
Education.''} \emph{Education and Information Technologies} 28 (4):
4221--41. \url{https://doi.org/10.1007/s10639-022-11316-w}.

\leavevmode\vadjust pre{\hypertarget{ref-nscai2021}{}}%
NSCAI. 2021. {``Final Report: The National Security Commission on
Artificial Intelligence.''} \url{https://www.nscai.gov/}.

\leavevmode\vadjust pre{\hypertarget{ref-oneil2017}{}}%
O'neil, Cathy. 2017. \emph{Weapons of Math Destruction: How Big Data
Increases Inequality and Threatens Democracy}. Crown.

\leavevmode\vadjust pre{\hypertarget{ref-oakley2019}{}}%
Oakley, Jacob G. 2019. {``Purple Teaming.''} In, 105--15. Berkeley, CA:
Apress. \url{https://doi.org/10.1007/978-1-4842-4309-1_8}.

\leavevmode\vadjust pre{\hypertarget{ref-oremus2023}{}}%
Oremus, Will. 2023. {``Meet the Hackers Who Are Trying to Make AI Go
Rogue.''}
\url{https://www.washingtonpost.com/technology/2023/08/08/ai-red-team-defcon/}.

\leavevmode\vadjust pre{\hypertarget{ref-pavlopoulos2020}{}}%
Pavlopoulos, John, Jeffrey Sorensen, Lucas Dixon, Nithum Thain, and Ion
Androutsopoulos. 2020. {``Toxicity Detection: Does Context Really
Matter?''} \url{https://doi.org/10.48550/ARXIV.2006.00998}.

\leavevmode\vadjust pre{\hypertarget{ref-peters2022}{}}%
Peters, Uwe. 2022. {``Algorithmic Political Bias in Artificial
Intelligence Systems.''} \emph{Philosophy \& Technology} 35 (2): 25.
\url{https://doi.org/10.1007/s13347-022-00512-8}.

\leavevmode\vadjust pre{\hypertarget{ref-raji2019}{}}%
Raji, Inioluwa Deborah, and Joy Buolamwini. 2019. {``AIES '19: AAAI/ACM
Conference on AI, Ethics, and Society.''} In, 429--35. Honolulu HI USA:
ACM. \url{https://doi.org/10.1145/3306618.3314244}.

\leavevmode\vadjust pre{\hypertarget{ref-raji2020}{}}%
Raji, Inioluwa Deborah, Andrew Smart, Rebecca N. White, Margaret
Mitchell, Timnit Gebru, Ben Hutchinson, Jamila Smith-Loud, Daniel
Theron, and Parker Barnes. 2020. {``FAT* '20: Conference on Fairness,
Accountability, and Transparency.''} In, 33--44. Barcelona Spain: ACM.
\url{https://doi.org/10.1145/3351095.3372873}.

\leavevmode\vadjust pre{\hypertarget{ref-ren2023}{}}%
Ren, Xiaozhe, Pingyi Zhou, Xinfan Meng, Xinjing Huang, Yadao Wang,
Weichao Wang, Pengfei Li, et al. 2023. {``PanGu-Sigma: Towards Trillion
Parameter Language Model with Sparse Heterogeneous Computing.''}

\leavevmode\vadjust pre{\hypertarget{ref-rosenblatt1958}{}}%
Rosenblatt, F. 1958. {``The Perceptron: A Probabilistic Model for
Information Storage and Organization in the Brain.''}
\emph{Psychological Review} 65 (6): 386--408.
\url{https://doi.org/10.1037/h0042519}.

\leavevmode\vadjust pre{\hypertarget{ref-schmidt2021}{}}%
Schmidt, Eric, Bob Work, Safra Catz, Steve Chien, Chris Darby, Kenneth
Ford, Jose-Marie Griffiths, et al. 2021. {``National Security Commission
on Artificial Intelligence (Ai).''} \emph{National Security Commission
on Artificial Intellegence, Tech. Rep}.

\leavevmode\vadjust pre{\hypertarget{ref-silver2016}{}}%
Silver, David, Aja Huang, Chris J. Maddison, Arthur Guez, Laurent Sifre,
George Van Den Driessche, Julian Schrittwieser, et al. 2016.
{``Mastering the Game of Go with Deep Neural Networks and Tree
Search.''} \emph{Nature} 529 (7587): 484--89.
\url{https://doi.org/10.1038/nature16961}.

\leavevmode\vadjust pre{\hypertarget{ref-smith2022}{}}%
Smith, Shaden, Mostofa Patwary, Brandon Norick, Patrick LeGresley,
Samyam Rajbhandari, Jared Casper, Zhun Liu, et al. 2022. {``Using
Deepspeed and Megatron to Train Megatron-Turing Nlg 530b, a Large-Scale
Generative Language Model.''} \emph{arXiv Preprint arXiv:2201.11990}.

\leavevmode\vadjust pre{\hypertarget{ref-soares2015}{}}%
Soares, Nate, and Benja Fallenstein. 2015. {``Aligning Superintelligence
with Human Interests: A Technical Research Agenda.''} In.
\url{https://api.semanticscholar.org/CorpusID:14393270}.

\leavevmode\vadjust pre{\hypertarget{ref-szegedy2013}{}}%
Szegedy, Christian, Wojciech Zaremba, Ilya Sutskever, Joan Bruna,
Dumitru Erhan, Ian Goodfellow, and Rob Fergus. 2013. {``Intriguing
Properties of Neural Networks.''}
\url{https://doi.org/10.48550/ARXIV.1312.6199}.

\leavevmode\vadjust pre{\hypertarget{ref-titus2023}{}}%
Titus, Alexander J. 2023. {``Violet Teaming AI in the Life Sciences,''}
July. \url{https://doi.org/10.5281/ZENODO.8180395}.

\leavevmode\vadjust pre{\hypertarget{ref-titus2018}{}}%
Titus, Alexander J., Audrey Flower, Patrick Hagerty, Paul Gamble,
Charlie Lewis, Todd Stavish, Kevin P. O'Connell, Greg Shipley, and
Stephanie M. Rogers. 2018. {``SIG-DB: Leveraging Homomorphic Encryption
to Securely Interrogate Privately Held Genomic Databases.''} \emph{PLOS
Computational Biology} 14 (9): e1006454.
\url{https://doi.org/10.1371/journal.pcbi.1006454}.

\leavevmode\vadjust pre{\hypertarget{ref-titus2023a}{}}%
Titus, Alexander J., Kathryn E. Hamilton, and Michelle Holko. 2023.
{``Cyber and Information Security in the Bioeconomy.''} In, 17--36.
Springer International Publishing.
\url{https://doi.org/10.1007/978-3-031-26034-6_3}.

\leavevmode\vadjust pre{\hypertarget{ref-unesco2021}{}}%
UNESCO, C. 2021. {``Recommendation on the Ethics of Artificial
Intelligence.''}

\leavevmode\vadjust pre{\hypertarget{ref-urbina2022}{}}%
Urbina, Fabio, Filippa Lentzos, Cédric Invernizzi, and Sean Ekins. 2022.
{``Dual Use of Artificial-Intelligence-Powered Drug Discovery.''}
\emph{Nature Machine Intelligence} 4 (3): 189--91.
\url{https://doi.org/10.1038/s42256-022-00465-9}.

\leavevmode\vadjust pre{\hypertarget{ref-vaswani2017}{}}%
Vaswani, Ashish, Noam Shazeer, Niki Parmar, Jakob Uszkoreit, Llion
Jones, Aidan N. Gomez, Lukasz Kaiser, and Illia Polosukhin. 2017.
{``Attention Is All You Need.''}
\url{https://doi.org/10.48550/ARXIV.1706.03762}.

\leavevmode\vadjust pre{\hypertarget{ref-wang2018}{}}%
Wang, Cliff, and Zhuo Lu. 2018. {``Cyber Deception: Overview and the
Road Ahead.''} \emph{IEEE Security \& Privacy} 16 (2): 8085.

\leavevmode\vadjust pre{\hypertarget{ref-whittaker2018}{}}%
Whittaker, Meredith, Kate Crawford, Roel Dobbe, Genevieve Fried,
Elizabeth Kaziunas, Varoon Mathur, Sarah Mysers West, et al. 2018.
\emph{AI Now Report 2018}. AI Now Institute at New York University New
York.

\leavevmode\vadjust pre{\hypertarget{ref-whittlestone2019}{}}%
Whittlestone, Jess, Rune Nyrup, Anna Alexandrova, and Stephen Cave.
2019. {``AIES '19: AAAI/ACM Conference on AI, Ethics, and Society.''}
In, 195--200. Honolulu HI USA: ACM.
\url{https://doi.org/10.1145/3306618.3314289}.

\leavevmode\vadjust pre{\hypertarget{ref-willdouglasheaven2022}{}}%
Will Douglas Heaven. 2022. {``Why Meta{'}s Latest Large Language Model
Survived Only Three Days Online.''}
\url{https://www.technologyreview.com/2022/11/18/1063487/meta-large-language-model-ai-only-survived-three-days-gpt-3-science/}.

\leavevmode\vadjust pre{\hypertarget{ref-winfield2019a}{}}%
Winfield, Alan. 2019. {``Ethical Standards in Robotics and AI.''}
\emph{Nature Electronics} 2 (2): 46--48.
\url{https://doi.org/10.1038/s41928-019-0213-6}.

\leavevmode\vadjust pre{\hypertarget{ref-winfield2019}{}}%
Winfield, Alan F., Katina Michael, Jeremy Pitt, and Vanessa Evers. 2019.
{``Machine Ethics: The Design and Governance of Ethical AI and
Autonomous Systems {[}Scanning the Issue{]}.''} \emph{Proceedings of the
IEEE} 107 (3): 509--17.
\url{https://doi.org/10.1109/JPROC.2019.2900622}.

\leavevmode\vadjust pre{\hypertarget{ref-yeung2020}{}}%
Yeung, Karen. 2020. {``Recommendation of the Council on Artificial
Intelligence (OECD).''} \emph{International Legal Materials} 59 (1):
2734.

\leavevmode\vadjust pre{\hypertarget{ref-zeide2022}{}}%
Zeide, Elana. 2022. {``The Silicon Ceiling: How Artificial Intelligence
Constructs an Invisible Barrier to Opportunity.''} \emph{UMKC L. Rev.}
91: 403.

\leavevmode\vadjust pre{\hypertarget{ref-zenko2015}{}}%
Zenko, Micah. 2015. \emph{Red Team: How to Succeed by Thinking Like the
Enemy}. Basic Books.

\leavevmode\vadjust pre{\hypertarget{ref-zheng2023}{}}%
Zheng, Ou, Mohamed Abdel-Aty, Dongdong Wang, Zijin Wang, and Shengxuan
Ding. 2023. {``ChatGPT Is on the Horizon: Could a Large Language Model
Be All We Need for Intelligent Transportation?''}

\end{CSLReferences}

\bibliographystyle{unsrt}
\bibliography{references.bib}

\end{document}